# Optimal Wildfire Escape Route Planning for Drones under Dynamic Fire and Smoke

Chang Liu, *Member, IEEE,* Tamas Sziranyi, *Senior Member, IEEE*

*Abstract*—In recent years, the increasing prevalence and intensity of wildfires have posed significant challenges to emergency response teams. The utilization of unmanned aerial vehicles (UAVs), commonly known as drones, has shown promise in aiding wildfire management efforts. This work focuses on the development of an optimal wildfire escape route planning system specifically designed for drones, considering dynamic fire and smoke models. First, the location of the source of the wildfire can be well located by information fusion between UAV and satellite, and the road conditions in the vicinity of the fire can be assessed and analyzed using multi-channel remote sensing data. Second, the road network can be extracted and segmented in real time using UAV vision technology, and each road in the road network map can be given priority based on the results of road condition classification. Third, the spread model of dynamic fires calculates the new location of the fire source based on the fire intensity, wind speed and direction, and the radius increases as the wildfire spreads. Smoke is generated around the fire source to create a visual representation of a burning fire. Finally, based on the improved *A\** algorithm, which considers all the above factors, the UAV can quickly plan an escape route based on the starting and destination locations that avoid the location of the fire source and the area where it is spreading. By considering dynamic fire and smoke models, the proposed system enhances the safety and efficiency of drone operations in wildfire environments.

*Keywords*— unmanned aerial vehicles (UAVs), Sentinel-2, optimal routes planning, weighted *A\** algorithm，wildfire rescue

## I. INTRODUCTION

In recent years, the frequent occurrence of wildland fires has posed an unprecedented and enormous challenge to humanity [1], especially wildland mountain fires that last up to several months and even destroy human homes [2]. As the modern world grapples with the consequences of increasingly severe wildfires, the utilization of UAVs has emerged as a promising avenue for enhancing the safety and efficacy of wildfire management [3]. Drones are potentially advantageous in responding to natural disasters because of their flexibility, such as providing the task of sending life jackets in the event of human drowning, among others [4]. It is well known that sentinel-2 satellites can detect the occurrence of wildfires, and generally forest fires last for a long period of time [5], so after the satellite detects the occurrence of forest fires, it is necessary for human beings to take timely measures to minimize the damage.

In this work, we propose the technique of fusion of 13-channel image information from Sentinel-2 satellite imagery and real-time image information captured by UAVs, and the cooperation between the two can accomplish the rescue mission in the event of wildfires more efficiently. First, the Sentinel-2 satellite can provide the UAVs system with information about the wildfire in advance, while the weather information of the area can also be pre-downloaded, and this step accomplishes wildfire detection and localization. Secondly, UAVs, due to their flexibility, can quickly reach the wildfire burning areas that are inaccessible to rescuers, search for humans in need of help through UAV vision technology, and utilize the real-time video sequences captured by them for the extraction of the road network, as well as the assessment of the road conditions of the road network in the area, where our previous works [6,7] have been well-prepared. Third, based on the road-map network information and road condition information extracted from the UAV in the previous step, dynamic fire and smoke spread models are added in this work, which affects the optimal path planning algorithm to go and change the best path in real time to avoid the location of the fire source and smoke, and to choose the best escape route with the best road condition for the people who are escaping.

In the subsequent sections, Section 2 describes the related work that has been done. Section 3 describes the overall architecture of the proposed system, the dynamic fire and smoke models, and the algorithms for dynamic optimal escape route planning. In Section 4, we demonstrate the experimental comparison results with a case study of a mountain wildfire that occurred on August 24, 2022, in Chongqing. Finally, we summarize and outlook this work.

## II. RELATED WORKS

### A. Wildfire Detection and Localization by Drone and Satellite Fusion Technology

Figure 1 illustrates wildfire detection and localization accomplished through the collaboration of Sentinel-2 satellites and UAVs. The figure is partially derived from the results of our previous work [8]. Sentinel-2 satellite carries a Multi-Spectral Imager (MSI) with a swath of 290 km. It provides a versatile set of 13 spectral bands spanning from the visible and near infrared to the shortwave infrared, featuring four bands at 10 m, six bands at 20 m and three bands at 60 m spatial resolution [9]. In full operational phase, the pair of S2 satellites will deliver data taken over all land

Chang Liu is with the Department of Networked Systems and Services, Budapest University of Technology and Economics, Műegyetem rkp. 3, H-1111 Budapest, Hungary and Machine Perception Research Laboratory of HUN-REN Institute for Computer Science and Control (SZTAKI), H-1111 Budapest, Kende u. 13-17, Hungary (corresponding author to provide phone: +36-30-526-4122; e-mail: liu.chang@sztaki.hun-ren.hu).

Tamas Sziranyi is with the Faculty of Transportation Engineering and Vehicle Engineering, Budapest University of Technology and Economics (BME-KJK), Műegyetem rkp. 3, H-1111 Budapest, Hungary and Machine Perception Research Laboratory of HUN-REN Institute for Computer Science and Control (SZTAKI), H-1111 Budapest, Kende u. 13-17, Hungary (e-mail: sziranyi.tamas@sztaki.hun-ren.hu).

surfaces and coastal zones every five days under cloud-free conditions, and typically every 15–30 days considering the presence of clouds [10]. From the wildfire information provided by the satellite data, the location of the fire source and the direction of the smoke can be clearly confirmed by using a combination of different bands. In the satellite image on the upper left in Figure 1, the red pixels indicate the location of the fire source, and the blue pixels show the direction of the smoke [8]. UAVs can quickly fly to the wildfire site after receiving the information, and due to the long duration of the mountain fires, timely rescue has been necessary for measures such as the mountain fire spreading blockage.

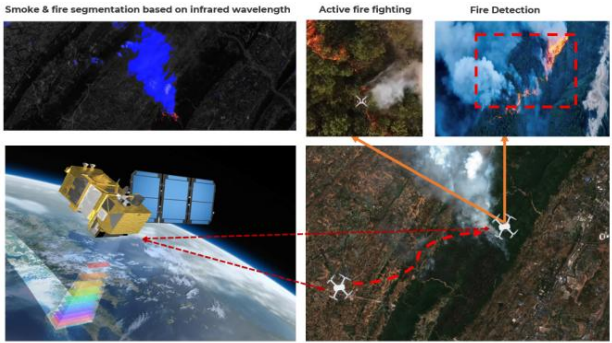

Fig. 1. Wildfire detection and localization through the cooperation of Sentinel-2 and UAVs.

### B. Roadmap Extraction and Road Surface Conditions Evaluation

After the drone arrives at the wildfire site, the drone's on-board GPU processes the real-time video sequences captured by its camera. Firstly, by segmenting the video sequences into image sequences. And after that, the images are delivered to the *D-LinkNet* [11] for road network extraction, which has as input an RGB image with (1024 * 1024 * 3) resolution, and the output is the image of the road network, with the black color representing the background, and white pixels represent the road extraction information. At the same time, the drones also analyze and evaluate the road conditions in the current area [7], so as to give priority to the roads with different road conditions. The green pixels represent the roads with good road conditions, and the white pixels represent the roads with bad road conditions. The evaluation of the road conditions combines the road surface materials, the weather conditions, and other factors. Figure 2 from our previous work [8] shows the above process in detail.

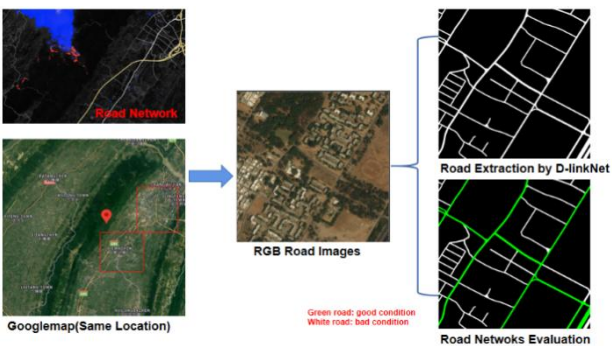

Fig. 2. Road Extraction and Road Condition Evaluation Result in the Wildfire [8].

### C. Rescue Gesture Recognition for People in Wildfire Distress

The drone can work with satellites to find the location of the fire source and the direction of the fire, and with its flexibility it can reach places that are inaccessible to rescuers and dangerous to them. Of course, it can also adjust its own flight altitude to perform different tasks. When the altitude of the drone is lowered, within 10 meters of the ground, the drone can search for humans in distress. Through human detection and the recognition of human dynamic rescue gestures [6], it can find people in need of help and provide them with help and escape routes. A related rescue gesture recognition dataset and the reliable model can be found in our previous work [6].

### III. METHODOLOGY

Based on the related work presented in Section 2, the main system framework of this work is shown in Figure 3. In Section 3 Methodology, we focus on the workflow of the whole system and what is different from the previous work. The dynamic fire and smoke spread model, which has not been considered before, is added to this work, and this dynamic model affects the working mechanism of the path planning algorithm in the previous work [12], and the modified algorithm will also be introduced at the end of this chapter.

### A. Framework of the Proposed System

Figure 3 is a flowchart of the entire system for dynamic escape path planning for wildfires. The pre-inputs to the system are the satellite image from sentinel-2 which is multi-channel and the weather information. Based on the 13-channel satellite image data processing, the center location of the fire source and the direction of the smoke will be well segmented and presented. At the same time, the system can also receive information about the weather in the vicinity of the wildfire during its occurrence. By fusing the information from the images captured by the UAV with the images provided by the satellite, the system can pinpoint the location of the wildfire.

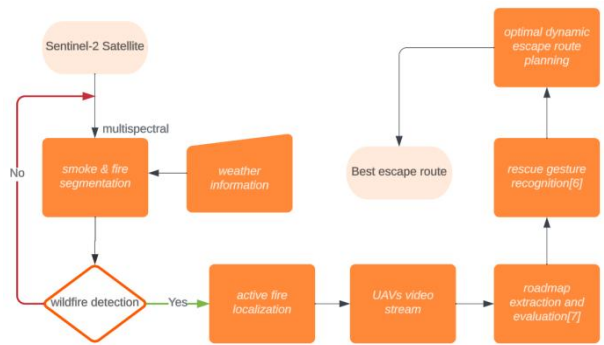

Fig. 3. Framework of the proposed wildfire escape route planning system.

Once the system detects and locates a wildfire, the drone can quickly fly to the location and begin rescue efforts. With the video stream captured by the drone at high altitude, the GPU-boarded drone can process the data in real time, segmenting the data into images that can be fed into the *D-LinkNet* [11] for road extraction. The extracted road network map, combined with the drone's assessment of road conditions, assigns road priority to the network, with road sections in good condition being given priority access, and vice versa for sections in poor condition [7]. This priority assignment takes into account factors such as weather information, road surface material, and road throughput.

After that, the UAV adjusts its flight altitude and descends from a high altitude to about 10 meters above the ground, the UAV starts detecting humans, recognizes the rescue gestures of the human after detecting a human being [6], and when it accurately recognizes a human needing help by dynamic attention gesture, the UAV starts performing the task of optimal escape route planning. Dynamic fire and smoke spread modeling is added to the above weighted road network, and the modified *A\** algorithm with weights provides an optimal escape route within a few seconds if the user provides the starting and ending points.

*B. Dynamic Fire Model and Smoke Model*

- Dynamic Fire Model: It calculates the new position of the fire source based on the wind direction and speed, which represents the dynamic nature of the fire. The wind direction is updated by the weather information, introducing variability in fire behavior. Wind direction can significantly affect how a fire spreads. The radius of the fire increases with each step, demonstrating the dynamic growth of the fire over time. A red circle is drawn at the position of the fire source, providing a visual representation of the fire's location. The fire's spread is determined probabilistically to simulate real-world fire dynamics. A binary mask is created to represent the neighboring cells that the fire can spread to. The resulting image shows simulated fire propagation, considering wind effects and fire spread.

- Dynamic Smoke Model: The smoke generation model generates random smoke particles around the fire source. Smoke intensity controls the number of smoke particles generated, representing the dynamic generation of smoke. Each smoke particle has a random gray intensity, giving the appearance of smoke. Smoke particles are added to the resulting image around the fire source, creating a dynamic smoke effect. The position and intensity of smoke particles are updated by the current weather information, adding variability to the smoke's appearance, which is characteristic of dynamic smoke models. The result image is updated to include the added smoke particles, creating a visual representation of a burning fire with dynamic smoke.

*C. Proposed Modified A\* Algorithm for Optimal Escape Route Planning under Dynamic Fire and Smoke Model*

Parallel to the fire and smoke spread modeling is the search for dynamic optimal paths, which we have modified on the *A\** algorithm [12]. The *A\** algorithm combines a heuristic search with a shortest path-based search [13]. It is defined as a best-first algorithm because each cell in the configuration space is evaluated by value:

$$f(n) = h(n) + g(n) \qquad (1)$$

where *g(n)* represents the cost from the starting point to the current node; *h(n)* represents the estimated cost from the current node to the ending point; n is the current node. *f(n)* is the total cost of the node.

In this work, among several heuristic mathematical functions *h(n)* we have chosen to use the $h_D$ to calculate the diagonal distance with the weighted modification [14]:

$$dx = |x_n - x_g| \qquad (2)$$

$$dy = |y_n - y_g| \qquad (3)$$

$$h_D(n) = d_1 * max(dx, dy) + (d_2 - d_1) * min(dx, dy) \qquad (4)$$

where $(x_n, y_n)$ is the coordinate of the current node n; $(x_g, y_g)$ is the coordinate of the end node n; For green cell, which are also pavements with good road conditions $d_1$=1 and $d_2$=1.4 (octile distance), white cell, which are also pavements with bad road conditions $d_1$=100 and $d_2$=140.

Unlike the path search algorithm in our previous work [12], the case of the *g(n)* function in this work has changed. The *g(n)* function in the previous search algorithm only needs to consider green pixel points and white pixel points. Black pixels are the background, and the algorithm cannot access the black pixel points. The green pixel points represent the roads with good road conditions which have priority, while the white pixel points represent the roads with poor road conditions which do not have priority. In this work, because of the need to account for dynamic fire and smoke spread, there is a risk of red fire pixel points and gray smoke pixel points appearing on roads with the green and white pixels. This makes it necessary for the optimal escape path search algorithm to consider these two scenarios that are dangerous to humans. As the fire spreads and the smoke spreads, the original path will be obstructed or interrupted, so we improved the previous algorithm [12] by categorizing it into two main cases, Figure 4 and Figure 5.

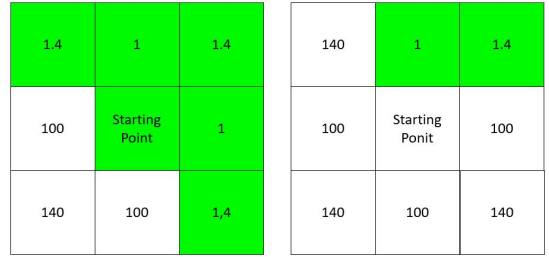

Case 1: Road with diffrenet conditions(Green road: good condition, white road: bad condition)

Fig. 4. Different assignment of *g(n)* cost value to different pixel points[12] (*Case 1*).

Case 1 shows how the *g(n)* function works without fire smoke dynamic propagation. In Fig. 4, the value of *g(n)* in the road network map is different when the user's starting position is on a road with different priority. The left plot of Fig. 4 demonstrates that when the user's starting position is on a road in good condition, the values of *g(n)* assigned to the eight neighbors of that pixel point are shown in the figure, and the value of the green pixel point is much smaller than the value of the white pixel point. The figure on the right side of Fig. 4 shows that when the user's starting position is on a road in poor condition, the values of the *g(n)* of the eight neighbors of the pixel are assigned as shown in the figure, and the values of the eight neighbors around the pixel are based on whether the road is in good condition or not and whether it has the priority to pass or not.

Fig. 5. Different assignment of *g(n)* cost value to different pixel points (Case 2).

Case 2 shows how the *g(n)* function works under the influence of the dynamic propagation of fire and smoke. In Figure 5, the *g(n)* values in the road network map differ when the user's starting position is located on a road with different priorities. Both cases illustrated in Fig. 5 are affected by dynamic fire and smoke propagation, as can be seen by the fact that the red fire pixel point is surrounded by gray smoke pixels. It is well known that in this case, even if the road conditions at that pixel point are good, it is forbidden for the user to pass through, because the smoke area and the fire source area are very dangerous for the pedestrians or vehicles on the road.

The left side of Fig. 5 shows a case where the user starting position is on a green pixel point with a good road condition, but there is a threat of fire and smoke around it, so in this case the algorithm is not allowed to access the gray and red pixel points. The right side of Figure 5 illustrates another case where the user starting position is on a white pixel point with a poor road condition, and there is also a threat of fire and smoke around, so in this case the algorithm is also not allowed to access the gray and red pixel points and can only select the best white pixel point for path searching. This is significantly different from our previous work in that the improved *A\** algorithm takes into account the dynamic fire spread model as well as the smoke dispersion model to influence the planning of the optimal paths.

## IV. EXPERIMENTAL RESULTS

This work presents the results using the Jinyun Mountain wildfire that occurred in Chongqing, China on August 24, 2022, as a case study. This section focuses on showing how the modified *A\** algorithm can find the optimal escape routes in the case of dynamic fire and smoke spread. We control the number of fire sources in the same area, and Figures 6 to 8 show the comparative results of path planning in the case of two three as well as four fire sources in turn. The roads consisting of green pixels in the three sets of comparison figures indicate roads with priority, i.e., good road conditions, and the roads consisting of white pixels represent roads in poor condition and without priority. The blue circle represents the departure location, and the yellow circle represents the destination. The red line is the result of optimal path planning. The red area is the fire source area and the gray area near the fire source indicates smoke.

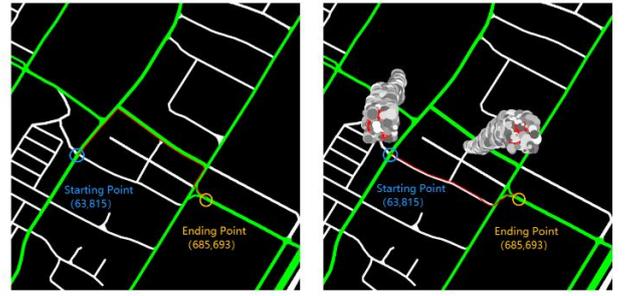

Fig. 6. Dynamic optimal escape route planning result comparison (2 fire sources).

Figure 6 demonstrates a comparison of the results of the first set of experiments, where there are two fire and smoke threats under that escape path planning, and the spread of fire and smoke can be seen in the right-hand panel of Figure 6. Both have the same starting point location (63,815) and ending point location (685,693). As can be seen in Fig. 6, when there is no dynamic threat of fire as well as smoke, the escape route found by the best path algorithm is to use as many roads as possible with good road conditions, and at this time, the *f-value* obtained is also relatively small at 587.2. When there is interference from fire as well as smoke, the optimal escape path planning algorithm selects the roads with white pixels, i.e., the roads in bad condition, because fire and smoke block the roads during the path planning process over time. At this point the value of *f* becomes large, up to 24411.2.

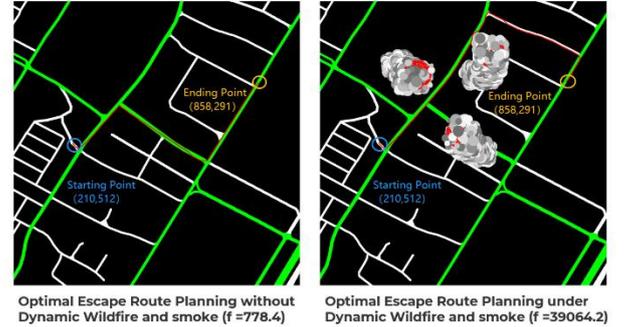

Fig. 7. Dynamic optimal escape route planning result comparison (3 fire sources).

Figure 7 demonstrates a comparison of the results of the second set of experiments, where there are three fire and smoke threats under that escape path planning, and the spread of fire and smoke can be seen in the right-hand panel of Figure 7. Both have the same starting point location (210,512) and ending point location (858,291). From Fig. 7, when there is no dynamic threat of fire as well as smoke, the optimal path algorithm finds an escape route that uses as many roads as possible that are in good condition, and at this point the *f-value* obtained is also relatively small at 587.2. When there is interference from fire and smoke, the optimal escape path planning algorithm also applies as many green pixels of roads in good condition as possible and selects as few white pixels of roads in poor condition as possible. The difference between the two paths can be seen on the right-hand side of Fig. 7 where the spread of the three fires as well as the smoke blocks many of the roads in good condition. The final *f value* with fire as well as smoke interference is as high as 39064.2.

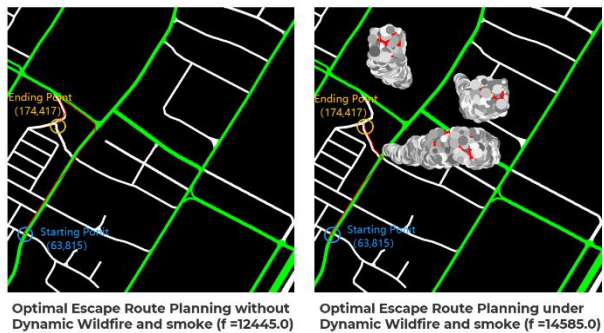

Fig. 8. Dynamic optimal escape route planning result comparison (4 fire sources).

Figure 8 illustrates a comparison of the results of the third set of experiments where there are four fire and smoke threats under this escape path planning. As can be seen in Figure 8, the four fire sources as well as the smoke over time two of them ended up merging together, which is why there is a large fire source and smoke in Figure 8. The user's starting location (63, 851) and ending location (174, 417) are the same for that scenario. In the absence of fire and smoke threats, the optimal path search algorithm uses as many roads in good condition as possible and avoids roads in poor condition. When there is a threat of fire and smoke, it can be seen from Fig. 8 that the merged two fire sources block a road, which affects the search results of the algorithm, and the best path is finally found as shown in Fig 8. By comparing the *f-values*, the *f-values* in these two cases are 12445 and 14585 respectively.

## V. CONCLUSION

This work presents an innovative approach to wildfire escape route planning for drones and demonstrates the versatility of drones in assessing road conditions, detecting people in distress, recognizing rescue gestures, and ultimately planning escape routes. This multi-functional approach leverages the full potential of UAVs in disaster response scenarios. The main notable strength in this work is incorporation of a dynamic fire and smoke propagation model, which takes into account various factors, providing real-time updates for optimal escape route planning.

The system proposed in this work, firstly, detects the location of fire source by fusing the information from satellite remote sensing images and UAV images. Secondly, the UAV accomplishes the tasks of road network extraction and road surface condition assessment by adjusting the flight altitude, and in low altitude the UAV accomplishes the detection of people in distress and the recognition of dynamic rescue gestures, which lays the foundation for best path planning. Finally, the dynamic fire and smoke propagation model is added to the weighted path search algorithm to provide a real-time optimal escape route based on the starting positions and ending positions provided by the pedestrians on the road. The results confirm that our proposed algorithm is effective. We validate our approach by applying it to a real wildfire case study. This practical application reinforces the relevance and effectiveness of our system.

This work is based on the Jinyun Mountain fire that occurred in Chongqing, China on August 24, 2022, as a case study, and we will test more cases and a broader range of wildfire scenarios and environments where wildfires occur in the future. We will also compare the results of more path search algorithms. Last but not least, more factors will be considered into the fire and smoke spread models, such as the geomorphology, as well as the vegetation index of the various surrounding areas and so on.

ACKNOWLEDGMENT

The work was carried out at HUN-REN Institute for Computer Science and Control, Hungary and was funded by the Hungarian National Science Foundation (NKFIH OTKA) No. K139485. The authors would like to thank their student Morui Zhu for the Chongqing wildfire Sentinel-2 dataset. The research work was also supported by project no. 2022-2.1.1-NL-2022-00012 National Laboratory of Cooperative Technologies is funded by the Ministry of Culture and Innovation through the National Research, Development and Innovation Fund, under the funding of the 2022-2.1.1-NL Establishment and Complex Development of National Laboratories call, by the European Union within the framework of the National Laboratory for Autonomous Systems (RRF-2.3.1–21-2022–00002) programs, and by the TKP2021-NVA-01 project.